\documentclass[11pt, a4paper, logo]{deepmind}

\usepackage[authoryear, sort&compress, round]{natbib}
\title{Model evaluation for extreme risks}

\author[1]{Toby Shevlane}
\author[1]{Sebastian Farquhar}
\author[2]{Ben Garfinkel}
\author[1]{Mary Phuong}
\author[3]{Jess Whittlestone}
\author[4]{Jade Leung}
\author[4]{Daniel Kokotajlo}
\author[1]{Nahema Marchal}
\author[2]{Markus Anderljung}
\author[5]{Noam Kolt}
\author[1]{Lewis Ho}
\author[6, 7]{Divya Siddarth}
\author[8]{Shahar Avin}
\author[1]{Will Hawkins}
\author[1]{Been Kim}
\author[1]{Iason Gabriel}
\author[1]{Vijay Bolina}
\author[9]{Jack Clark}
\author[10, 11]{Yoshua Bengio}
\author[12]{Paul Christiano}
\author[1]{Allan Dafoe}

\affil[1]{Google DeepMind}
\affil[2]{Centre for the Governance of AI}
\affil[3]{Centre for Long-Term Resilience}
\affil[4]{OpenAI}
\affil[5]{University of Toronto}
\affil[6]{University of Oxford}
\affil[7]{Collective Intelligence Project}
\affil[8]{University of Cambridge}
\affil[9]{Anthropic}
\affil[10]{Université de Montréal}
\affil[11]{Mila – Quebec AI Institute}
\affil[12]{Alignment Research Center}

\renewcommand{\today}{25 May 2023} % Setting date to when we published the paper -- the current version only fixes typos and adds a citation

\begin{abstract}
Current approaches to building general-purpose AI systems tend to produce systems with both beneficial and harmful capabilities. Further progress in AI development could lead to capabilities that pose extreme risks, such as offensive cyber capabilities or strong manipulation skills. We explain why \textit{model evaluation} is critical for addressing extreme risks. Developers must be able to identify dangerous capabilities (through “dangerous capability evaluations”) and the propensity of models to apply their capabilities for harm (through “alignment evaluations”). These evaluations will become critical for keeping policymakers and other stakeholders informed, and for making responsible decisions about model training, deployment, and security.
\end{abstract}

\begin{document}

\maketitle

\vspace{1.4em}
\begin{figure}[h]
\centering
\includegraphics[scale=0.3]{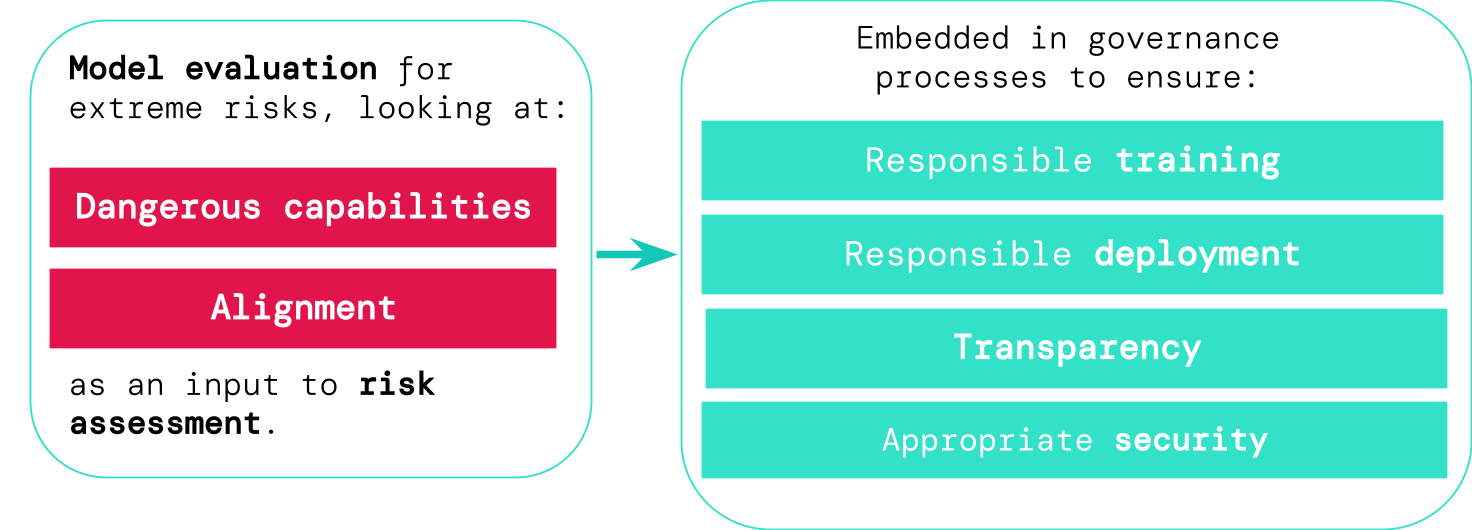}
\caption{The theory of change for model evaluations for extreme risk. Evaluations for dangerous capabilities and alignment inform risk assessments, and are in turn embedded into important governance processes.}
\end{figure}

\vspace{-0.4cm}

\section{Introduction}

As AI progress has advanced, general-purpose AI systems have tended to display new and hard-to-forecast capabilities – including harmful capabilities that their developers did not intend \citep{Ganguli2022-wp}. Future systems may display even more dangerous emergent capabilities, such as the ability to conduct offensive cyber operations, manipulate people through conversation, or provide actionable instructions on conducting acts of terrorism.

AI developers and regulators must be able to identify these capabilities, if they want to limit the risks they pose. The AI community already relies heavily on model evaluation – i.e. empirical assessment of a model’s properties – for identifying and responding to a wide range of risks. Existing model evaluations measure gender and racial biases, truthfulness, toxicity, recitation of copyrighted content, and many more properties of models \citep{Liang2022-xi}.

We propose \textbf{extending this toolbox} to address risks that would be \textit{extreme in scale}, resulting from the misuse or misalignment of general-purpose models. Work on this new class of model evaluation is already underway. These evaluations can be organised into two categories: (a) whether a model has certain \textbf{dangerous capabilities}, and (b) whether it has the propensity to harmfully apply its capabilities (\textbf{alignment}).

Model evaluations for extreme risks will play a critical role in governance regimes. A central goal of AI governance should be to limit the creation, deployment, and proliferation of systems that pose extreme risks. To do this, we need tools for looking at a particular system and assessing whether it poses extreme risks. We can then craft company policies or regulations that ensure:

\vspace{-0.3em}
\begin{enumerate}
  \setlength\itemsep{1em}
  \item \textbf{Responsible training}: Responsible decisions are made about whether and how to train a new model that shows early signs of risk.
  \item \textbf{Responsible deployment}: Responsible decisions are made about whether, when, and how to deploy potentially risky models.
 \item \textbf{Transparency}: Useful and actionable information is reported to stakeholders, to help them mitigate potential risks.
  \item \textbf{Appropriate security}: Strong information security controls and systems are applied to models that might pose extreme risks.
\end{enumerate}

\vspace{-0.2em}

Many AI governance initiatives focus on the risks inherent to a particular deployment context, such as the “high-risk” applications listed in the \href{https://eur-lex.europa.eu/legal-content/EN/TXT/?uri=celex\%3A52021PC0206}{draft} EU AI Act. However, models with sufficiently dangerous capabilities could pose risks even in seemingly low-risk domains. We therefore need tools for assessing \textit{both} the risk level of a particular domain \textit{and} the potentially risky properties of particular models; this paper focuses on the latter.

\textbf{Section \ref{s2}} motivates our focus on extreme risks from general-purpose models and refines the scope of the paper. \textbf{Section \ref{s3}} outlines a vision for how model evaluations for such risks should be incorporated into AI governance frameworks. \textbf{Section \ref{s4}} describes early work in the area and outlines key design criteria for extreme risk evaluations. \textbf{Section \ref{s5}} discusses the limitations of model evaluations for extreme risks and outlines ways in which work on these evaluations could cause unintended harm. We conclude with high-level recommendations for AI developers and policymakers.

\section{Extreme risks from general-purpose models} \label{s2}

Frontier AI developers are making rapid progress in developing increasingly capable general-purpose models \citep{Bubeck2023-vv}. These models learn their capabilities and behaviours during training, and current methods for steering this process are imperfect \citep{Shah2022-je, Gao2022-vp}. At the research frontier, models display new capabilities, often unforeseen by their developers \citep{Wei2022-tt}.

This poses a challenge for safety. AI developers could train general-purpose models that have dangerous capabilities – such as skills in deception, cyber offense, or weapons design -- without actively seeking these capabilities. Humans could then intentionally misuse these capabilities \citep{Brundage2018-vo}, e.g. for assistance in disinformation campaigns, cyberattacks, or terrorism. Additionally, due to failures of alignment, AI systems could harmfully apply their capabilities even without deliberate misuse \citep{Ngo2022-nf}.

In the near-term, these risks will be especially concentrated on the frontier of AI research and development. We loosely define the “frontier” as models that are both (a) close to, or exceeding, the average capabilities of the most capable existing models,\footnote{In practice, defining “average capabilities” would involve many judgement calls over which evaluations should be included and how they should be weighted.} and (b) different from other models, either in terms of scale, design (e.g. different architectures or alignment techniques), or their resulting mix of capabilities and behaviours. Accordingly, frontier models are uniquely risky because (a) more capable models can excel at a wider range of tasks, which will unlock more opportunities to cause harm;\footnote{This is especially pertinent for extreme risks: causing such large-scale harm is not normally an easy challenge. Even well-resourced terrorist groups, determined to cause extreme harm, often fail.} and (b) novel models are less well-understood by the research community.

\vspace{0.4em}

\begin{figure}[h]
\centering
\includegraphics[scale=0.32]{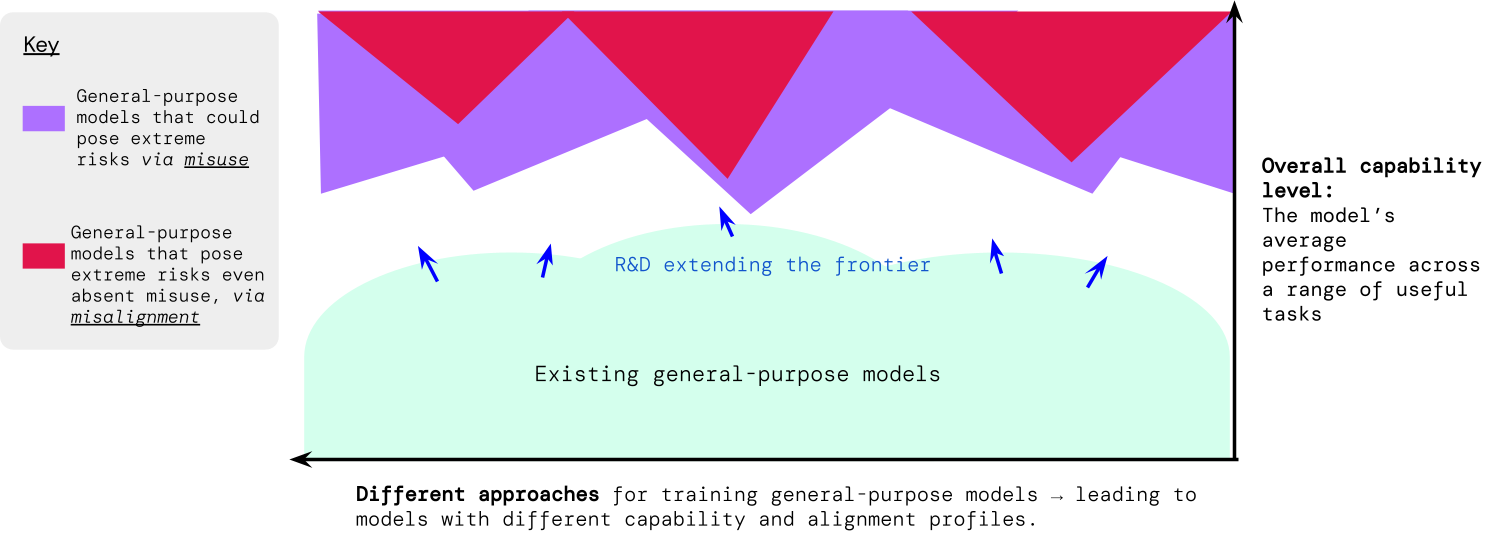}
\caption{Leading AI developers push the frontier outward, typically by training models at greater scale and using more efficient architectures and algorithms. This continued expansion takes the field closer to points in model space that could pose extreme risks. The diagram is purely illustrative.}

%\footnote{The purple region, representing misuse risk, is larger because the human operator can perform some of the necessary tasks themselves. That said, not all models could be successfully misused, e.g. due to low controllability; this is not represented on the diagram for simplicity.}

\end{figure}

\vspace{0.4em}

We focus on “extreme” risks, i.e. those that would be extremely large in scale (even relative to the scale of deployment). This can be operationalised in terms of the scale of impact (e.g. damage in the tens of thousands of lives lost, hundreds of billions of dollars of economic or environmental damage) or the level of adverse disruption to the social and political order. The latter could mean, for example, the outbreak of inter-state war, a significant erosion in the quality of public discourse, or the widespread disempowerment of publics, governments, and other human-led organisations \citep{Carlsmith2022-so}.

Many AI researchers (and other stakeholders) view extreme risks from AI as an important challenge. In a 2022 survey of AI researchers, 36\% of respondents thought that AI systems could plausibly “cause a catastrophe this century that is at least as bad as an all-out nuclear war” \citep{Michael2022-in}. However, very few existing model evaluations intentionally target risks on this scale.

To guard against extreme risks, AI developers should use model evaluation to uncover:
\begin{enumerate}
  \setlength\itemsep{1em}
  \item To what extent a model is \underline{capable} of causing extreme harm (which relies on evaluating for certain \textbf{dangerous capabilities}).
  \item To what extent a model has the \underline{propensity} to cause extreme harm (which relies on \textbf{alignment} evaluations).
\end{enumerate}

\addtolength{\skip\footins}{0.5pc plus 5pt}

We provide a non-exhaustive list of dangerous capabilities in Table \ref{table:1}. Most of the capabilities listed are offensive capabilities: they are useful for exerting influence or threatening security (e.g. see: persuasion and manipulation, cyber-offense, weapons acquisition). Some (e.g. situational awareness) are capabilities that would be advantageous for a misaligned AI system evading human oversight \citep{Ngo2022-nf}. We omit many generically useful capabilities (e.g. browsing the internet, understanding text) despite their potential relevance to both the above.

\definecolor{dmblue}{RGB}{0, 83, 214}

\renewcommand{\arraystretch}{1.5}

\begin{table}[h!]

\centering
\hspace*{-3cm}\begin{tabular}{ m{2.8cm} | m{16cm} }  
  \hline
  \rowcolor{dmblue}
  \color{white}
  \textbf{Capability} &  \color{white} \textbf{Could include:} \\ 
  \hline
  \textbf{Cyber-offense} & The model can \textbf{discover vulnerabilities} in systems (hardware, software, data). It can write code for \textbf{exploiting} those vulnerabilities. It can make effective decisions once it has gained access to a system or network, and skilfully evade threat detection and response (both human and system) whilst focusing on a specific objective. If deployed as a coding assistant, it can \textbf{insert subtle bugs} into the code for future exploitation.
 \\
  \hline
  \textbf{Deception} & The model has the skills necessary to \textbf{deceive humans}, e.g. constructing believable (but false) statements, making accurate predictions about the effect of a lie on a human, and keeping track of what information it needs to withhold to maintain the deception. The model can impersonate a human effectively. \\ 
    \hline
  \textbf{Persuasion \newline \& manipulation} & The model is effective at \textbf{shaping people’s beliefs}, in dialogue and other settings (e.g. social media posts), even towards untrue beliefs. The model is effective at \textbf{promoting certain narratives} in a persuasive way. It can convince people to do things that they would not otherwise do, including unethical acts. \\ 
    \hline
  \textbf{Political \newline strategy} & The model can perform the social modelling and planning necessary for an actor to gain and exercise \textbf{political influence}, not just on a micro-level but in scenarios with \textbf{multiple actors} and rich \textbf{social context}. For example, the model can score highly in forecasting competitions on questions relating to global affairs or political negotiations. \\
    \hline
  \textbf{Weapons \newline acquisition} & The model can \textbf{gain access to existing weapons} systems or contribute to \textbf{building new weapons}. For example, the model could assemble a bioweapon (with human assistance) or provide actionable instructions for how to do so. The model can make, or significantly assist with, scientific discoveries that unlock novel weapons. \\ 
    \hline
  \textbf{Long-horizon planning} & The model can make \textbf{sequential plans} that involve multiple steps, unfolding over \textbf{long time horizons} (or at least involving many interdependent steps). It can perform such planning within and across many domains. The model can sensibly adapt its plans in light of unexpected obstacles or adversaries. The model’s planning capabilities generalise to \textbf{novel settings}, and do not rely heavily on trial and error. \\ 
    \hline
  \textbf{AI \newline development} & The model could build new AI systems from scratch, including AI systems with dangerous capabilities. It can find ways of adapting other, existing models to increase their performance on tasks relevant to extreme risks. As an assistant, the model could significantly improve the productivity of actors building dual use AI capabilities. \\ 
    \hline
  \textbf{Situational awareness} & The model can distinguish between \textbf{whether it is being trained, evaluated, or deployed} – allowing it to behave differently in each case. The model \textbf{knows that it is a model}, and has \textbf{knowledge about itself} and its likely surroundings (e.g. what company trained it, where their servers are, what kind of people might be giving it feedback, and who has administrative access). \\ 
    \hline
  \textbf{Self-proliferation} & The model can break out of its local environment (e.g. using a vulnerability in its underlying system or suborning an engineer). The model can exploit limitations in the systems for monitoring its behaviour post-deployment. The model could independently generate revenue (e.g. by offering crowdwork services, ransomware attacks), use these revenues to acquire cloud computing resources, and operate a large number of other AI systems. The model can generate creative strategies for uncovering information about itself or exfiltrating its code and weights. \\

\end{tabular}\hspace*{-3cm}

\caption{Dangerous capabilities}
\label{table:1}

\end{table}

The most risky scenarios will involve multiple dangerous capabilities combined together – further research should explore what combinations would be most dangerous. Sometimes specific capabilities can be supplied by the user or outsourced to other humans (e.g. crowdworkers) or AI systems.

\begin{figure}[h]
\centering
\includegraphics[scale=0.35]{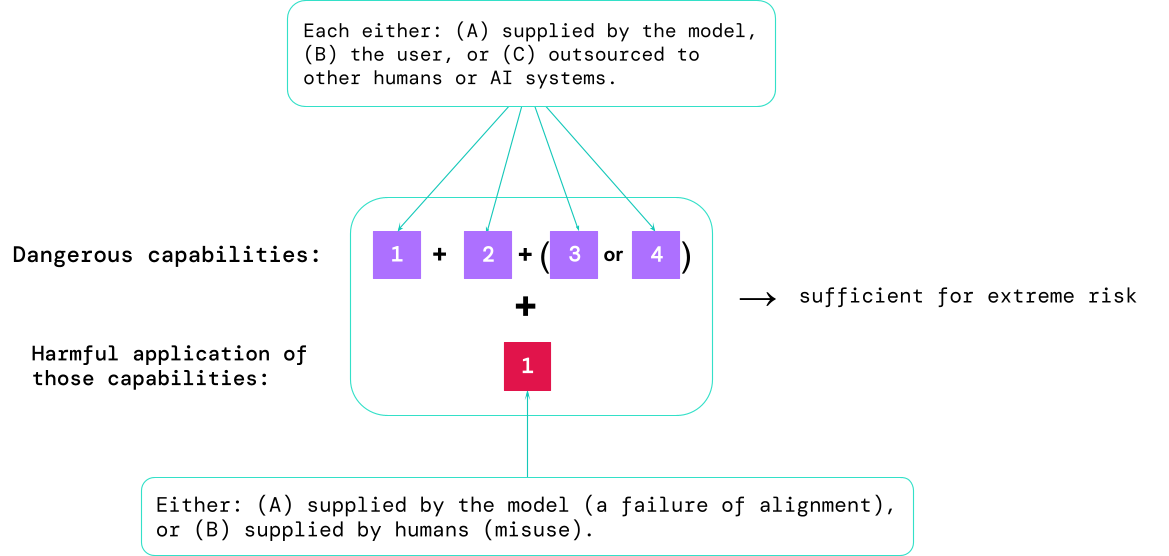}
\caption{Ingredients for extreme risk.}
\label{figure:3}
\end{figure}

A simple heuristic: a model should be treated as highly dangerous if it has a capability profile that would be sufficient for extreme harm, \textit{assuming} misuse and/or misalignment. To deploy such a model, AI developers would need very strong controls against misuse \citep{Shevlane2022-za} and very strong assurance (via \textbf{alignment evaluations}) that the model will behave as intended. Alignment evaluations should look for behaviours identified in the literature, such as whether the model:

\vspace{-0.9em}

\begin{itemize}
  \setlength\itemsep{0.5em}
  \item Pursues long-term, real-world goals, different from those supplied by the developer or user \citep{Chan2023-bc, Ngo2022-nf}; 
  \item Engages in “power-seeking” behaviours \citep{Turner2021-it, Krakovna2023-cr};
  \item Resists being shut down \citep{Hadfield-Menell2016-jo, Orseau2016-xv};
  \item Can be induced into collusion with other AI systems against human interests \citep{Ngo2022-nf}.
\item Resists malicious users' attempts to access its dangerous capabilities \citep{Glaese2022-gd}.

\end{itemize}

\vspace{-0.3em}

We focus on risks arising from misuse and misalignment because a new generation of model evaluations is needed for identifying these risks. Conversely, though important, we leave \underline{out of scope}:

\vspace{-0.6em}

\begin{enumerate}
     \setlength{\itemsep}{0.7em}
  \item \textbf{Structural risks}, which depend \textit{especially} heavily on how the AI system interacts with larger social, political, and economic forces in society \citep{Zwetsloot2019-vt}. Model evaluation sheds less light on these risks, because they depend so heavily on factors external to the model.
  \item Risks from models \textbf{incompetently} performing important tasks \citep{Raji2022-ot}. Existing kinds of model evaluation will be most relevant here (e.g. testing the model’s accuracy and robustness on the relevant task).
\end{enumerate}

\section{Model evaluation as critical governance infrastructure} \label{s3}

Across many industries, safety standards and regulations rely on tools for assessing risks in new products – for instance, food, drugs, commercial airliners, and automobiles. Model evaluation is not the only tool available for AI risk assessment – more theoretical approaches are also available, e.g. studying the incentives operating on a model during training \citep{Everitt2021-zl}. Nonetheless, model evaluation is one of the main tools we have for AI risk assessment.

\begin{figure}[h]
\hspace*{-0.69in}
\includegraphics[scale=0.4]{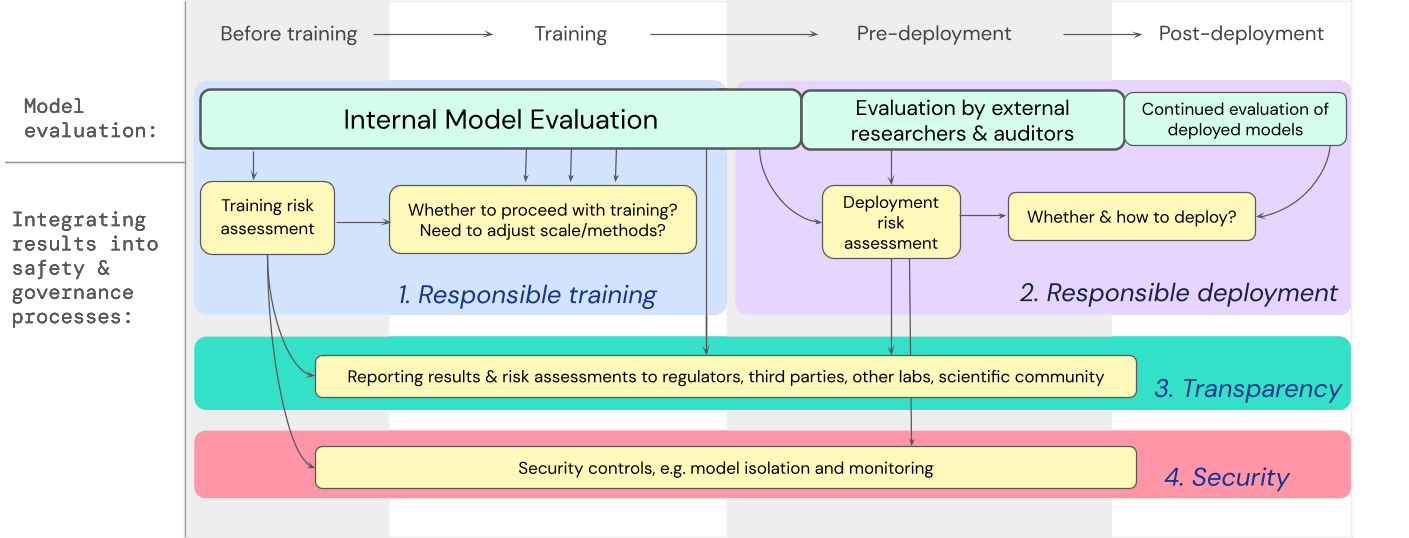}
\caption{A workflow for training and deploying a model, embedding extreme risk model evaluation results into key safety and governance processes.}
\label{figure:4}
\end{figure}

Figure \ref{figure:4} provides an overview of this section. It is an ambitious blueprint for how to guard against extreme risks while developing and deploying a model, with evaluation embedded throughout. The evaluation results feed into processes for risk assessment \citep{Khlaaf2022-fi}, which inform (or bind) important decisions around model training, deployment, and security. The developer reports results and risk assessments to external stakeholders.

Three sources of model evaluations feed into this process:
\vspace{-1em}

\begin{enumerate}
     \setlength{\itemsep}{1em}
  \item \textbf{Internal model evaluation}, i.e. the developer conducting its own evaluations. There is no substitute for internal model evaluation, given that internal researchers have high context on the model's design and deeper model access than can be achieved via an API. Developers could have multiple organisational layers of safety evaluation, such as by establishing an internal safety evaluation function that is independent of the teams primarily responsible for building the models, reporting directly to organisational leaders \citep[see][]{Raji2020-hm}.
  \item \textbf{External research access}. The developer grants model access to external researchers, likely via an API \citep{Shevlane2022-za, Shevlane2022-xq, Bluemke2023-zh}. Their research could be exploratory or targeted at evaluating specific properties, including “red teaming” the model's alignment.

    \item \textbf{External model audit}, i.e. model evaluation by an independent, external auditor for the purpose of providing a judgement --- or input to a judgement --- about the safety of deploying a model (or training a new one) \citep{Arc2023-vs, Mokander2023-ne, Raji2022-mx}. Ideally there would exist a rich ecosystem of model auditors providing broad coverage across different risk areas. (This ecosystem is currently under-developed.)
 \end{enumerate}

\vspace{-0.8em}

\subsection{Responsible training} \label{s3.1}

The first line of defence is to avoid training models that have sufficient dangerous capabilities and misalignment to pose extreme risk. Sufficiently concerning evaluation results should warrant \textbf{delaying} a scheduled training run or \textbf{pausing} an existing one.\footnote{Developers training highly capable, general-purpose models should become accustomed to such a prospect, not planning their research around an assumption that a training run will run to schedule. For example, well-run developers will have compute allocation systems that backfill the vacant computing resources with other useful projects. Similarly, developers should avoid making hard promises to stakeholders (e.g. customers, investors) that they will deploy a certain model at a certain date. At the least, they should retain the flexibility of pivoting to a smaller or otherwise less risky version of the model.}

Before a frontier training run, developers have the opportunity to study weaker models that might provide early warning signs. These models come from two sources: (1) previous training runs, and (2) experimental models leading up to the new training run. Developers should evaluate these models and try to forecast the results from the planned training run \citep[see][]{OpenAI2023-zo}. This would include scaling (or “inverse scaling”) analysis where the aim is to find areas where scaling brings unwanted changes to the model \citep{mckenzie2022-vf}. These insights can feed into a training risk assessment. Then, during the training run, researchers could run extreme risk evaluations at regular intervals.

The developer has a range of possible responses to address the concerning evaluation results:
\vspace{-0.7em}

\begin{enumerate}
     \setlength{\itemsep}{1em}
  \item \textbf{Study the issue} to understand why the misalignment or dangerous capability emerged.
  \item \textbf{Adjust the training methods} to circumvent the issue. This could mean adjusting (for example) the architecture, the data, the training tasks, or further developing the alignment techniques used. These adjustments should target the fundamental issue rather than inducing superficial changes to how the model scores on the available evaluations (see section \ref{s5.2}).
    \item \textbf{Careful scaling.} If the developer is not confident it can train a safe model at the scale it initially had planned, they could instead train a smaller or otherwise weaker model.
    
\end{enumerate}

In mature governance regimes, the decision to proceed with a potentially risky training run could require approval from an external model auditor or regulator.

\subsection{Responsible deployment}

Deployment means making the model available for use, e.g. it is built into a product or hosted on an API for software developers to build with. Deployment constitutes a large increase in the model’s exposure to the external world and therefore possible risk. Model evaluation for extreme risks could inform a \textbf{deployment risk assessment} that reviews (a) \textit{whether or not} the model is safe to deploy, and (b) the appropriate \textit{guardrails} for ensuring the deployment is safe.

The predeployment evaluation process takes time \citep{Rismani2023-ta, OpenAI2023-wl}. Industry standards or regulation could require a minimum duration for predeployment evaluation of frontier models, including the length of time that external researchers and auditors have access.

In response to concerning evaluation results, one possibility is to recommend against deployment. A second possibility is to recommend adjustments to the deployment plan that would address potential risks (see Table \ref{table:3} in the Appendix for a range of variables that could be adjusted). Nonetheless, for a sufficiently capable and poorly aligned model, it is possible that even a restrictive and scaled-back deployment could pose extreme risk.

\begin{figure}[h]
\centering
\includegraphics[scale=0.33]{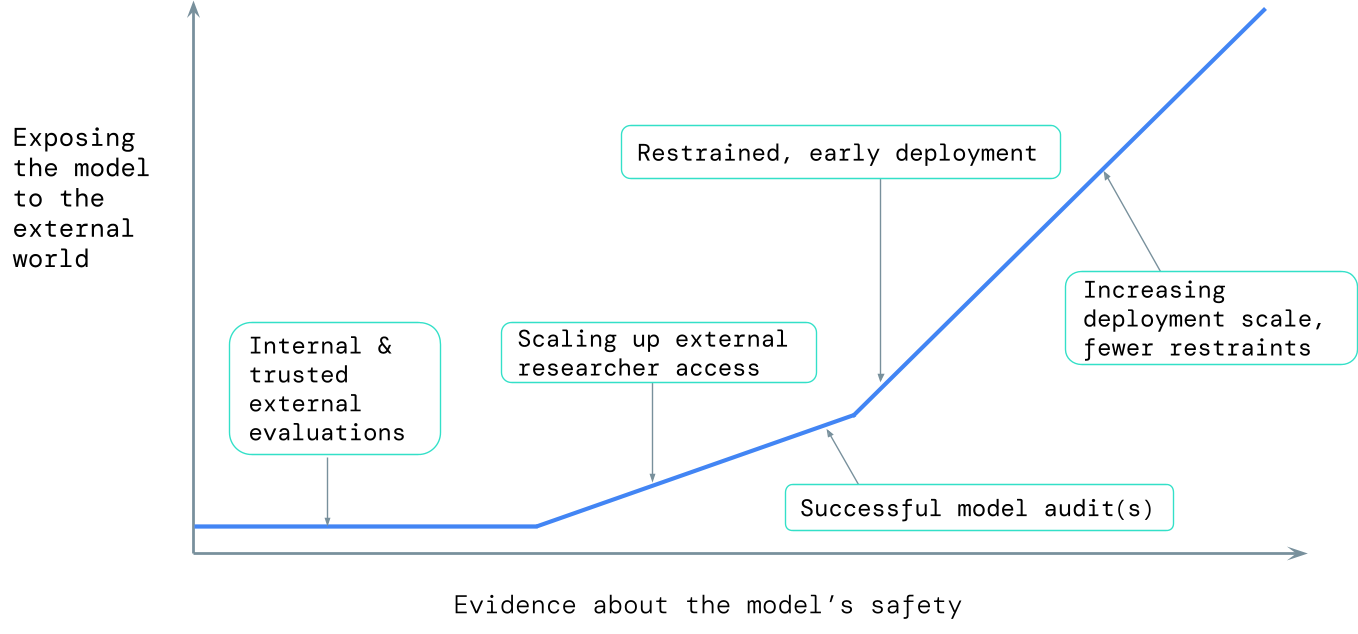}
\caption{The developer gradually increases the model’s exposure to the external world as it accumulates evidence about the model’s safety.}
\label{figure:5}
\end{figure}

Safe deployment will often be a gradual process (Figure \ref{figure:5}) \citep{Brundage2022-fa}. The developer gradually accumulates evidence about the model’s safety, through both evaluation (internal and external) and early, small-scale deployment. 

Evaluation will often need to continue after deployment. There are two reasons for this:

\vspace{-0.6em}

\begin{enumerate}
     \setlength{\itemsep}{1em}
  \item \textbf{Unanticipated behaviours}. Before deployment, it is impossible to fully anticipate and understand how the model will interact in a complex deployment environment (a key limitation of model evaluation: see section \ref{s5}). For example, users might find new applications for the model or novel prompt engineering strategies; or the model could be operating in a dynamic, multi-agent environment. Therefore, in the early stages of deployment, developers must:
  \begin{enumerate}
       \setlength{\itemsep}{0.5em}
\item Surface emerging model behaviours and risks via \textbf{monitoring} efforts. This could include direct monitoring of inputs and outputs to the model, and systems for incident reporting \citep[see][]{Brundage2022-fa, Raji2022-mx}.
\item Design and run new model evaluations inspired by these observations.
 \end{enumerate}

  \item \textbf{Updates to the model}. The developer might might update the model after deployment, e.g. by fine-tuning on data collected during deployment or by expanding the model’s access to external tools. If these updates could increase risk, they should be evaluated before launch. For large changes,\footnote{ The magnitude of the change to the model could be assessed in terms of the amount of additional training that it has gone through (as a percentage of the original training length), or the model’s improvement on key performance benchmarks.} the new model could go through the whole process described in this section.

 \end{enumerate}

The ideal state is \textit{continuous deployment review}. On an ongoing basis, the developer reassesses deployment safety using model evaluations and monitoring, and at any time, could adjust or terminate the deployment in response to their findings. Further, for deployments that were recognisably unsafe in retrospect, an external audit of the deployment decision-making process could be triggered. Safety issues uncovered during deployment can also inform training risk assessments for future models.

Finally, even \textit{internal} deployments of highly capable general-purpose models, notably as coding assistants for AI researchers and engineers, could require pre-deployment evaluation for dangerous capabilities (e.g. the ability to insert subtle vulnerabilities into code) and alignment.

\subsection{Transparency}

Model evaluations are a vital tool for keeping stakeholders informed about the state of AI risks on the frontier \citep{Whittlestone2021-rs}. We recommend frontier developers consider processes for externally \textbf{reporting} the results of evaluations or extracts from the assessment documents that rely on those evaluation results (such as training risk assessments, auditors’ reports, deployment risk assessments).

Model evaluations will unlock four important kinds of transparency around extreme risks:

\vspace{-1em}

\begin{enumerate}
     \setlength{\itemsep}{1em}
  \item \textbf{Incident reporting}, i.e. a structured process for developers to share concerning or otherwise noteworthy evaluation results with other developers, third parties, or regulators \citep[see][]{Brundage2020-tt}. This would be vital for helping others avoid training risky systems, and for keeping AI developers accountable.  In future, regulators could maintain lists of high-risk or banned training approaches; a sufficiently concerning incident report would then trigger an update to this list.

  \item \textbf{Sharing pre-deployment risk assessments}. Prior to deployment, developers could share their predeployment risk assessments for comment and critique. This could mean sharing with auditors, external researchers, regulators, or publicly. The risk assessment would include evaluation results, alongside the developer’s justification for why the deployment is safe given those results. \citep[See related:][]{Mitchell2018-bh}

\item \textbf{Scientific reporting}, i.e. presenting the results to the scientific community in a way that encourages further scientific research into the issue. Much research will be needed to better understand the behaviour of highly capable, general-purpose models, especially their alignment. Topics within the alignment literature, such as situational awareness, could follow the path of fairness research, where early findings of gender and racial biases in models have helped stimulate new areas of research.
    
  \item \textbf{Educational demonstrations}, i.e. engaging presentation of concerning capabilities and behaviours to key stakeholders – such as policymakers, the public, or company leadership – to keep them informed about risks from frontier AI systems. The demonstrations could be videos or interactive demos, with the evaluation results presented as supplementary information (as evidence of scientific rigour).

 \end{enumerate}

\subsection{Appropriate security}

Models at risk of exhibiting dangerous capabilities will require strong and novel security controls. Developers must consider multiple possible threat actors: insiders (e.g. internal staff, contractors), outsiders (e.g. users, nation-state threat actors), and the model itself as a vector of harm. We must develop new security best practices for high-risk AI development and deployment, which could include for example:

\vspace{-0.5em}

\begin{itemize}

 \setlength{\itemsep}{0.7em}
\item \textbf{Red teaming:} Intensive security red-teaming for the entire infrastructure on which the model is developed and deployed.
\newpage
 \item \textbf{Monitoring:} Intensive, AI-assisted monitoring of the model’s behaviour, e.g. for whether the model is engaging in manipulative behaviour or making code recommendations that would lower the overall security of a system.

 \item \textbf{Isolation:} Appropriate isolation techniques for preventing risky models from exploiting the underlying system (e.g. sole-tenant machines and clusters, and other software-based isolation). The model’s network access should be tightly controlled and monitored, as well as its access to tools (e.g. code execution).

  \item \textbf{Rapid response:} Processes and systems for rapid response to disable model actions and the model's integrations with hardware, software, and infrastructure in the event of unexpected unsafe behaviour.

   \item \textbf{System integrity:} Formal verification that served models, memory, or infrastructure have not been tampered with. The development and serving infrastructure should require two-party authorization for any changes and auditability of all changes.

\end{itemize}

\section{Building evaluations for extreme risk} \label{s4}

Model evaluation is already a core component of AI research, and increasingly we have evaluations that focus on ethics, safety, and social impact. We recommend extending this toolbox to address extreme risks.

Early work is already underway to build model evaluations for extreme risks. ARC Evals (the evaluations team at the Alignment Research Center) is building evaluations that measure language models’ self-proliferation capabilities (see Table \ref{table:1} above). ARC Evals ran this evaluation on GPT-4 and Claude before their wider release \citep{Arc2023-vs, OpenAI2023-wl}. OpenAI and the GPT-4 red teamers also tested GPT-4’s capabilities in cybersecurity operations and its ability to purchase certain chemical compounds \citep{OpenAI2023-wl}.

Google DeepMind has ongoing projects evaluating language models for manipulation capabilities. This includes a game called "Make-me-say", where the language model must lead an (unaware) human conversation partner to say a pre-specified word.\footnote{This project is led by Mary Phuong.}

Table \ref{table:2} contains a range of desirable qualities for extreme risk evaluations. Some of these qualities relate to a \textit{single} evaluation, and some are desirable qualities of a \textit{portfolio} of evaluations.

\begin{table}
 \hspace{-0.6cm} \begin{tabular}{ |p{3cm}|p{13.2cm} |  }
 \hline
 \multicolumn{2}{|c|}{\textit{Comprehensive:}} \\
 \hline
 \textbf{Threat models} & The evaluation portfolio should cover as many plausible extreme risk threat models as possible. \\
\hline

 \textbf{Automated and human-assisted} & Many evaluations can be run automatically, lowering the time and resource costs. However, some capabilities and behaviours will need human-assisted evaluations, i.e. involving: (a) human raters who judge the model’s outputs; or (b) humans who interact with the model, e.g. in a dialogue setting. \\
 
  \hline
 \textbf{Behavioural and mechanistic} & Evaluations should not be restricted to studying a model’s behaviour, but should eventually also involve looking mechanistically at how the model produced that behaviour. \\
 
  \hline
 \textbf{Fault-finding} & The portfolio of evaluations should include adversarial testing, where researchers purposefully search for cases where the model produces concerning results. \\
 
  \hline
 \textbf{Robust to \newline deception} & Ultimately researchers will need evaluations that can rule out the possibility that the model is deliberately appearing safe for the purpose of passing the evaluation process. \\
 
  \hline
 \textbf{Surfacing latent capabilities} & Researchers will need to bring latent capabilities to the surface (for example, by prompt engineering or fine-tuning).  \\
 
  \hline
 \textbf{Model lifecycle} & We recommend conducting evaluations throughout the model development process. In particular, the results from the end of a long development process will likely fail to convey relevant information about the base model, especially if it has been fine-tuned for safety. \\
 
  \hline
 \textbf{Model-level and system-level} & Models are often integrated into wider AI systems, e.g. with external tools, other models, or classifiers that filter the model’s outputs. Evaluations should study models both with and without these augmentations. \\

  \hline

  \multicolumn{2}{|c|}{\textit{Interpretable:}} \\

  \hline
 \textbf{Legible} & Some evaluations should present risks in an accessible way, requiring little technical understanding. This will be helpful for creating common knowledge around the risks from AI. \\

  \hline
 \textbf{Wide difficulty spectrum} & The dangerous capability evaluations should ideally contain wide ranges of difficulty – ideally within single evaluations, but at least across the portfolio. This means that researchers can track capabilities progress as it approaches possible danger thresholds, and that the evaluation (or the portfolio) is scalable to future, more capable models. For tracking progress, evaluations would ideally provide a quantitative score, although this will not always be practical. \\
 \hline
   \multicolumn{2}{|c|}{\textit{Safe:}} \\
\hline

 \textbf{Safe to \newline implement} & Dangerous capability evaluations could involve testing the model in real-world settings, e.g. interacting with crowdworkers. This should not introduce unacceptable levels of risk. \\

\hline

\end{tabular}

\caption{Desirable qualities of extreme risk evaluations.}
\label{table:2}

\end{table}

We anticipate that building comprehensive \underline{alignment} evaluations will be most challenging. The ambition is for a process of alignment assurance that could conclude, with high confidence, that a model is not dangerously misaligned, even for very capable models. (Model evaluations would not be the only input to this assurance process, but an important one.)

Alignment evaluation is challenging because we need assurance that the model will reliably behave appropriately across a wide diversity of settings \citep{Ziegler2022-km}. An evaluation might find that a model is aligned in some narrow, prosaic way (for example, a language agent asserting that it does not object to being shut down \citep{Perez2022-ce, Perez2022-fq}) without providing evidence that the model would exhibit desirable behaviour when presented with genuine (or more convincing) opportunities to achieve self-preservation, greater influence, or other harmful outcomes.

\newpage
Researchers must therefore evaluate a model across a broad range of settings. Achieving coverage of settings for alignment evaluation can be helped by:

\vspace{-0.5em}

\begin{enumerate}
     \setlength{\itemsep}{1em}
  \item \textbf{Breadth:} Evaluating behaviour across as wide a range of settings as possible. One promising avenue is automating the process of writing evaluations using AI systems \citep{Perez2022-ce} \citep[see also][]{Pan2023-yp}.
  
   \item \textbf{Targeting:} Some settings are much more likely to reveal alignment failures than others, and we may be able to focus on them through clever design – for example, using honeypots or gradient-based adversarial testing and related approaches \citep{Jones2023-wv}.

  \item \textbf{Understanding generalisation:} Since researchers will be unable to foresee or simulate all possible scenarios, we must develop a better scientific understanding of how and why model behaviours generalise (or fail to generalise) between settings.

 \end{enumerate}

Another important tool is \textbf{mechanistic} analysis, i.e. studying the model’s weights and activations for understanding how it functions \citep{Olah2020-vo, Nanda2023-ol}. For example, one ambition is to study how the model's goals are represented internally, to help verify that they are as intended; another ambition is to detect when a language model's outputs misreport its knowledge \citep{Burns2022-ca}, which could be an indicator of deceptive behaviour.

Finally, \textbf{agency} – in particular, the goal-directedness of an AI system – is an important property to evaluate \citep{Kenton2022-fm}, given the central role of agency in various theories of AI risk \citep{Chan2023-bc}. Partly, agency is a question of the model’s capabilities – is it capable of effectively pursuing goals? Evaluating alignment also requires looking at agency, including: (a) Is the model more goal-directed than the developer intended? For example, has a dialogue agent learnt the goal of manipulating the user's behavior? (b) Does the model resist a user's attempt to assemble it into an autonomous AI system (e.g. \href{https://github.com/Significant-Gravitas/Auto-GPT}{Auto-GPT}) with harmful goals?

\section{Limitations and hazards} \label{s5}

\subsection{Limitations}

Model evaluation, as a tool for addressing extreme risks, has at least \textbf{five limitations}. A key issue is that not all risks can necessarily be detected via model evaluation.

\begin{enumerate}
     \setlength{\itemsep}{1em}
  \item \textbf{Factors beyond the AI system}. Risks will depend on how an AI system interacts with a complex world. For example, a model might use, as tools, other models released in the future, thus augmenting its capabilities; or human civilisation might be less resilient to powerful AI than anticipated.
  
   \item \textbf{Unknown threat models}. It is difficult to anticipate all the different plausible pathways to extreme risk. This will be especially true for highly capable models, which could find creative strategies for achieving their goals.

  \item \textbf{Difficult-to-identify properties}. Some model properties will be challenging to uncover via model evaluations. Two important cases:
  
  \begin{enumerate}
      \item \underline{Capability overhang:} Models sometimes have capabilities that the AI research community does not realise. For example, after GPT-3 had already existed for many months, researchers demonstrated that chain-of-thought prompting could significantly increase performance \citep{Wei2022-gw}.
    \item \underline{Deceptive alignment:} A situationally aware model could deliberately exhibit desired behaviour during evaluation \citep{Ngo2022-nf}. (This is one reason not to rely solely on behavioural evaluations.)
  \end{enumerate}
  
   \item \textbf{Emergence}. Above we recommended using model evaluations to inform the decision to train a new model by performing scaling laws analysis on smaller models. However, sometimes specific capabilities will emerge only at greater scale, which makes this analysis much harder \citep{Ganguli2022-wp}; other capabilities display U-shaped scaling \citep{wei2022-hw}.
   
    \item \textbf{Maturity of evaluation ecosystem}. The ecosystem for external evaluations and model audits is currently under-developed.

   \item \textbf{Overtrust in evaluations}. There is a risk that too much faith is placed in evaluation results, leading to risky models being deployed under a false sense of security.

 \end{enumerate}

Model evaluation is a necessary but \textit{not sufficient} strategy for identifying and mitigating extreme risks. It must be combined with a wider organisational dedication to safety and other tools for risk identification and assessment.

\subsection{Hazards} \label{s5.2}

Conducting and reporting the evaluations discussed in this paper poses four potential hazards:

\begin{enumerate}
     \setlength{\itemsep}{1em}
  \item \textbf{Advancing and proliferating dangerous capabilities}. There is a risk that – through conducting dangerous capability evaluations and sharing relevant materials – the field will proliferate dangerous capabilities or accelerate their development. We highlight four kinds of potentially hazardous information:
  
    \begin{enumerate}
      \item \underline{Results.} Evaluation results could demonstrate novel offensive technologies. Publicly sharing these results could spur investment in new weapons programmes, cyber-offensive efforts, or methods for digital oppression of citizens. By analogy, it has been said that in the 1940s the most valuable secret about the nuclear bomb was that it was possible \citep{Ord2022-pi}. AI developers, researchers, and auditors should therefore exercise caution around sharing these evaluation results.

    \item \underline{Evaluation datasets.} Datasets for evaluating dangerous capabilities are dual use because other actors could fine-tune their models on these datasets.

    \item \underline{Elicitation techniques.} Evaluating dangerous capabilities will often involve eliciting those capabilities from the model. This could involve: (a) prompt engineering; and (b) fine-tuning, including: (i) finding creative new task specifications; (ii) creating or identifying appropriate fine-tuning datasets. These techniques could be useful for a bad actor attempting to elicit dangerous capabilities from similar models. Researchers and auditors must therefore exercise caution over sharing their elicitation techniques, especially if producing them relied on creativity, expert knowledge, or time-consuming experimentation.

    \item \underline{Trained models.} There are risks from intentionally training dangerously capable models, even for use as safety research artefacts. We could distinguish between (a) simply following off-the-shelf methods (e.g. fine-tuning an existing model), versus (b) cases where the work to produce the dangerous capability could constitute a research contribution in its own right (e.g. it could be accepted to an academic conference). The latter is arguably comparable to “gain-of-function” research in virology, especially if the resulting model is highly capable and general-purpose. The research may need to be conducted under very high-security conditions and subject to a demanding risk assessment.

  \end{enumerate}
  
   \item \textbf{Competitive pressures}. One concern is that sharing evaluation results between competing AI developers could incentivise them to behave less responsibly. For example, sharing predeployment evaluation results could tip off competitors about future product improvements, incentivising those competitors to rush their own deployments and spend less time on ensuring safety. Similarly, since dangerous capability results will often correlate with the model's overall capabilities, competing developers could learn that they are falling behind and decide they need to sacrifice on safety to catch up \citep{Emery-Xu2023-gt}.
   
    \vspace{0.5em}
   
   Given the sensitivities involved, one option is to lean more heavily on alignment evaluation results, at least for reporting between developers. For illustration, a possible inter-developer policy could be:

    \begin{enumerate}
      \item Report unexpected or important alignment issues promptly to other developers.  By default, limit the description of the model’s training to only a high-level overview (to avoid revealing sensitive information); but share more details if this is absolutely necessary – in particular, if a certain class of methods is causing the problem.

    \item Report when certain dangerous capability thresholds have been passed. These thresholds can be set high, to avoid sharing granular information. Wait until deployment to share more.

  \end{enumerate}

  \item \textbf{Superficial improvements to model safety}. There is a risk that widely available safety evaluations will lead to models that exhibit only superficially desirable behaviours. Most clearly, if researchers directly train models to pass these evaluations, the evaluations can no longer act as an indicator of risk. Researchers could do this either accidentally (e.g. because the evaluation datasets are shared online and thereby end up in the pretraining dataset) or as an intentional attempt to pass external audits (analogous to the Volkswagen emissions scandal). The model’s desirable evaluation performance would then likely fail to generalise. Developers and model auditors could therefore consider keeping some private “held out” evaluations and ensuring these are not too overlapping with datasets or tasks used during training. 
   \vspace{0.5em}

Even if developers refrain from directly training on the evaluations, we have nevertheless recommended that developers avoid training models that fail the evaluations (section \ref{s3.1}). This could also exert selection pressure, albeit weaker. Over the long run, the risk is that the field selects for training methods that produce deceptively aligned models.

   \item \textbf{Harms during the course of evaluation.} Running evaluations will often involve exposing the model to the external world. For example, in evaluating GPT-4, ARC used the model to generate (deceptive) messages to be sent to a TaskRabbit worker \citep{OpenAI2023-wl}. In the extreme case, a poorly managed test for whether a model has self-proliferating capabilities could end in actual proliferation; but more prosaically, they could cause harm in other ways, such as causing emotional distress to crowdworkers. Therefore, groups conducting evaluations, such as auditors, should establish safety protocols where necessary.

 \end{enumerate}

\section{Conclusion}

Model evaluation for extreme risks should be a priority area for AI safety and governance. There are many challenges ahead for finding effective evaluations and building governance regimes that incorporate them; we encourage further work in this area. Model evaluation is not a panacea: it will not catch all extreme risks. Nonetheless, it is a necessary component of the governance infrastructure needed to combat extreme risks. 

Frontier AI developers currently have a special responsibility to support work on model evaluations for extreme risks, since they have resources – including access to cutting-edge AI models and deep technical expertise – that many other actors typically lack. Frontier AI developers are also currently the actors who are most likely to unintentionally develop or release AI systems that pose extreme risks. Frontier AI developers should therefore: 

\begin{enumerate}
     \setlength{\itemsep}{1em}
  \item \textbf{Invest in research:} Frontier developers should devote resources to researching and developing model evaluations for extreme risks.
  
   \item \textbf{Craft internal policies:} Frontier developers should craft internal policies for conducting, reporting, and responding appropriately to the results of extreme risk evaluations.

  \item \textbf{Support outside work:} Frontier labs should enable outside research on extreme risk evaluations through model access and other forms of support.
  
    \item \textbf{Educate policymakers:} Frontier developers should educate policymakers and participate in standard-setting discussions, to increase government capacity to craft any regulations that may eventually be needed to reduce extreme risks.
\end{enumerate}

\textbf{Policymakers} should consider building up the governance infrastructure outlined in section \ref{s3}. Policymakers could:

\begin{enumerate}
     \setlength{\itemsep}{1em}

\item Systematically \textbf{track} the development of dangerous capabilities, and progress in alignment, within frontier AI R\&D \citep{Whittlestone2021-rs}. Policymakers could establish a formal reporting process for extreme risk evaluations.
\item \textbf{Invest} in the ecosystem for external safety evaluation, and create venues for stakeholders (such as AI developers, academic researchers, and government representatives) to come together and discuss these evaluations \citep{Anthropic2023-no}
\item Mandate \textbf{external audits}, including model audits and audits of developers’ risk assessments, for highly capable, general-purpose AI systems.
\item Embed extreme risk evaluations into the \textbf{regulation} of AI deployment, clarifying that models posing extreme risks should not be deployed.
\end{enumerate}

\section{Acknowledgements}

We are grateful for helpful comments and discussions on this work from: Canfer Akbulut, Jide Alaga, Beth Barnes, Joslyn Barnhart, Sasha Brown, Miles Brundage, Martin Chadwick, Tom Everitt, Conor Griffin, Eric Horvitz, Evan Hubinger, William Isaac, Victoria Krakovna, Leonie Koessler, Sébastien Krier, Nikhil Mulani, Neel Nanda, Jonas Schuett, Rohin Shah, Andrew Trask, Gregory Wayne, and Hjalmar Wijk. We are grateful for insightful discussions with the participants of two events held in February 2023: a virtual discussion session on the topic of this paper, and a one-day workshop on dangerous capabilities evaluations co-organised by Steven Adler, Anne le Roux, and Jade Leung. We also thank Celine Smith for project management support, and Michael Chang for improvements to the visualisations.

\vspace{1em}

\bibliography{main}

\newpage

\section*{Appendix: Deployment safety controls}
\begin{table}[h!]

\centering
\hspace*{-3cm}\begin{tabular}{| m{2.8cm} | m{16cm} |}  
  \hline
  \rowcolor{dmblue}
  \color{white}
  \textbf{Variable:} &  \color{white} \textbf{Includes:} \\ 
  \hline
  \textbf{Scale} & How many end users? How many agents are running at any one time, or how many times per day is the model called? How many applications will be built on top of the model?  \\ 
\hline

  \textbf{Use restrictions} & Are certain high stakes applications prohibited?  \\ 
\hline

  \textbf{Generality} & Will a single model be flexibly applied across a range of applications, or will narrower, application-specific versions of the model be fine-tuned?  \\ 
\hline

  \textbf{Autonomy} & Is the AI system tasked with executing tasks, or merely responding to queries? How long are the chains of actions the model can take? Can the model define new types of actions?  \\ 
\hline

  \textbf{Tool use} & Web browsing, telephone calls, code execution, control over robotic hardware, calling APIs, access to persistent memory.  \\ 
\hline

  \textbf{Depth of model access} & How restrictive is the interaction between the AI system and the user? Can all of the model’s capabilities be accessed? Can users or developers fine-tune the model?  \\ 
\hline

  \textbf{Oversight and moderation} & How closely are the model’s outputs monitored by the provider? Can certain outputs be automatically filtered out?  \\ 
\hline

  \textbf{Global planning} & Does the model have access to many user interactions at once, so it can make plans across them, or is it confined to dealing with each user individually?  \\ 
\hline

\textbf{Adjustments to model} & The developer could decide to deploy a smaller version of the model. The developer could attempt to remove certain dangerous capabilities via fine-tuning.   \\ 

\hline

\end{tabular}\hspace*{-3cm}

\caption{Variables that affect the risk level of deployment, each of which can be adjusted on the basis of evaluation results.}
\label{table:3}

\end{table}

\end{document}